\documentclass[twoside,11pt]{article}
\usepackage{amsmath}
\usepackage{blindtext}
\usepackage{graphicx}
\usepackage{subcaption}
\usepackage{booktabs}
\usepackage{multirow}

\usepackage{jmlr2e}

\usepackage{lastpage}

\firstpageno{1}

\begin{document}

\title{Computationally-efficient deep learning models for nowcasting of precipitation: A solution for the Weather4cast 2025 challenge}


\author{\name Anushree Bhuskute \email anushree.bhuskute@flame.edu.in \\
       \addr Center for Interdisciplinary Artificial Intelligence\\
       Flame University\\
       Pune, Maharashtra 412115, India
       \AND
       \name Kaushik Gopalan \email kaushik.gopalan@flame.edu.in \\
       \addr Center for Interdisciplinary Artificial Intelligence (CAI) \& School of Computing and Data Sciences\\
       Flame University\\
       Pune, Maharashtra 412115, India
       \AND
       \name Jeet Shah \email  jeet.c.shah@flame.edu.in \\
       \addr Center for Interdisciplinary Artificial Intelligence\\
       Flame University\\
       Pune, Maharashtra 412115, India }
\maketitle

\begin{abstract}
This study presents a transfer-learning framework based on Convolutional Gated Recurrent Units (ConvGRU) for short-term rainfall prediction in the Weather4Cast 2025 competition. A single SEVIRI infrared channel ($10.8 \mu$m wavelength) is used as input, which consists of four observations over a one-hour period. A two-stage training strategy is applied to generate rainfall estimates up to four hours ahead. In the first stage, ConvGRU is trained to forecast the brightness temperatures from SEVIRI, enabling the model to capture relevant spatiotemporal patterns. In the second stage, an empirically derived nonlinear transformation maps the predicted fields to OPERA-compatible rainfall rates.

For the event-prediction task, the transformed rainfall forecasts are processed using 3D event detection followed by spatiotemporal feature extraction to identify and characterize precipitation events. Our submission achieved 2nd place in the cumulative rainfall task. Further, the same model was used out-of-the-box for the event prediction task, and resulted in similar scores as the baseline model to the competition. 
\end{abstract}

\begin{keywords}
  Precipitation nowcasting, ConvGRU, transfer learning, satellite-to-radar translation, extreme weather, probabilistic forecasting
\end{keywords}

\section{Introduction}

Short term precipitation forecasting, also known as nowcasting, plays a crucial role in reducing the impact of severe weather events. Precise and timely predictions enable timely flood warnings, optimize agriculture irrigation scheduling, support aviation safety protocols, and many more (see \cite{reichstein:2019}). Traditional numerical weather prediction (NWP) models are grounded in physical principles but struggle to capture rapidly evolving convective phenomena due to their coarse temporal resolution and high computational cost (see \cite{bauer_thorpe_brunet:2015,schultz_betancourt_gong_kleinert_langguth_leufen:2019}). 
Existing systems for operational precipitation nowcasting have been using optical flow for forecasting (see \cite{woo_wong:2017}). In the recent days, many data driven and hybrid machine learning and deep learning methods have emerged as efficient alternative for learning complex atmospheric patterns directly from observational data. Models such as CNN, RNN, U-Net have been used for precipitation nowcasting (see \cite{veillette_samsi_mattioli:2020}). PredRNN and ConvCNN help improve spatiotemporal detection for nowcasting (see \cite{shi:2015, wang_wu_zhang_gao_wang_yu_long:2022}).

The Weather4Cast competition series at NeurIPS provides a large-scale benchmark for satellite-based precipitation forecasting. Early editions (see \cite{gruca2023weather4cast}) focused on predicting high-resolution rainfall maps from SEVIRI satellite imagery under spatial and temporal shifts. Later versions (see \cite{deshpande:2024, leinonen2021spatiotemporal}) introduced probabilistic forecasting and emphasized uncertainty estimation, domain adaptation, and physical consistency. These challenges have established Weather4Cast as a key platform for evaluating deep learning models in spatiotemporal weather prediction. 

Precipitation nowcasting from satellite observations involves several challenges. First, it is a cross-domain problem requiring effective transfer learning from the satellite observation domain (radiance-based measurements) to the radar domain (precipitation rates). The two domains capture related but distinct atmospheric processes, and models must learn representations that generalize across this gap (see \cite{bauer_thorpe_brunet:2015}).
Second, the task demands spatiotemporal sequence modeling to represent the nonlinear evolution of atmospheric systems, including advection, growth, and dissipation of convective clouds. Third, probabilistic forecasting is required to quantify uncertainty over multi-hour horizons, as atmospheric chaos amplifies prediction variance with time (see \cite{rasp_dueben_scher_weyn_mouatadid_thuerey:2020}). Moreover, computational efficiency is vital for real-time deployment; models must balance expressiveness and speed. Finally, extreme event detection involves identifying rare high-intensity precipitation episodes while filtering noise across multiple spatial and temporal scales.

We propose a two-stage transfer learning pipeline using ConvGRU networks for efficient satellite-based rainfall forecasting. In Stage 1, ConvGRU models are trained to predict future SEVIRI brightness temperature fields from past observations, learning atmospheric dynamics directly in the satellite observation space. In Stage 2, an empirically calibrated transformation translates these predictions into rainfall estimates in the radar domain.

\section{Methodology}

The Weather4cast 2025 competition at NeurIPS focuses on evaluating generalization performance and emergent capabilities of probabilistic precipitation models through downstream application tasks. The competition provides SEVIRI satellite imagery as input and OPERA radar rainfall estimates as ground truth targets. The OPERA estimates, having 6 times the spatial resolution of SEVIRI, are resampled to SEVIRI's native 252×252 pixel resolution to create matching target images.  The challenge comprises two prediction tasks: (1) cumulative rainfall forecasting over 4-hour horizons, and (2) extreme precipitation event detection. (including event location, spatial-temporal extent, and severity). Our approach utilizes SEVIRI satellite imagery to predict future rainfall patterns, bridging the gap between satellite radiance observations and ground-based radar measurements.

The satellite comprises of 11-band spectral images covering visible (VIS), water vapor (WV), and infrared (IR) bands from the SEVIRI instrument. Each 252x252 image represents a 15-minute period and pixels corresponding to spatial areas ranging from 4km at the equator to 12km in the northern latitudes. All 11 bands are not equally informative for rainfall estimation. The primary indicator of precipitaiton likelyhood is the presence of cold cloud-top temperatures. The channels in the infrared window at 10.8 \textmu m (IR\_108) and 12.0 \textmu m (IR\_120) represent the most reliable information about cloud-top temperatures. These infrared channels have high correlation, thus we selected only 1 channel at 10.8 \textmu m (IR\_108) (see \cite{deshpande:2024}).

The original 252 x 252 input image was padded to 256 x 256 using constant zero padding to ensure compatibility with the model architecture's feature extraction operations and maintain efficient tensor operations on GPU hardware. 
For pre-processing, input brightness temperatures were normalized to the range [0, 1] by dividing the pixel values by 300, which was approximately equal to the maximum observed values in the training dataset. To enhance the model's ability to discriminate significant cloud structures from clear-sky backgrounds, we applied Otsu's adaptive thresholding method (see \cite{otsu:1975}) where we set the pixel values of the non cloudy regions as unity. This helped the model focus on relevant features and simultaneously reduced noise. Training was conducted using two NVIDIA RTX 6000 Ada Generation GPUs, with a total training time of about six hours. The code repository for the described process is available on \hyperlink{https://github.com/flame-cai/w4c25_rainfall}{GitHub} at \url{https://github.com/flame-cai/w4c25_rainfall}

\subsection{Model Architecture}
\begin{figure}[htbp]
    \centering
    \includegraphics[width=1\textwidth]{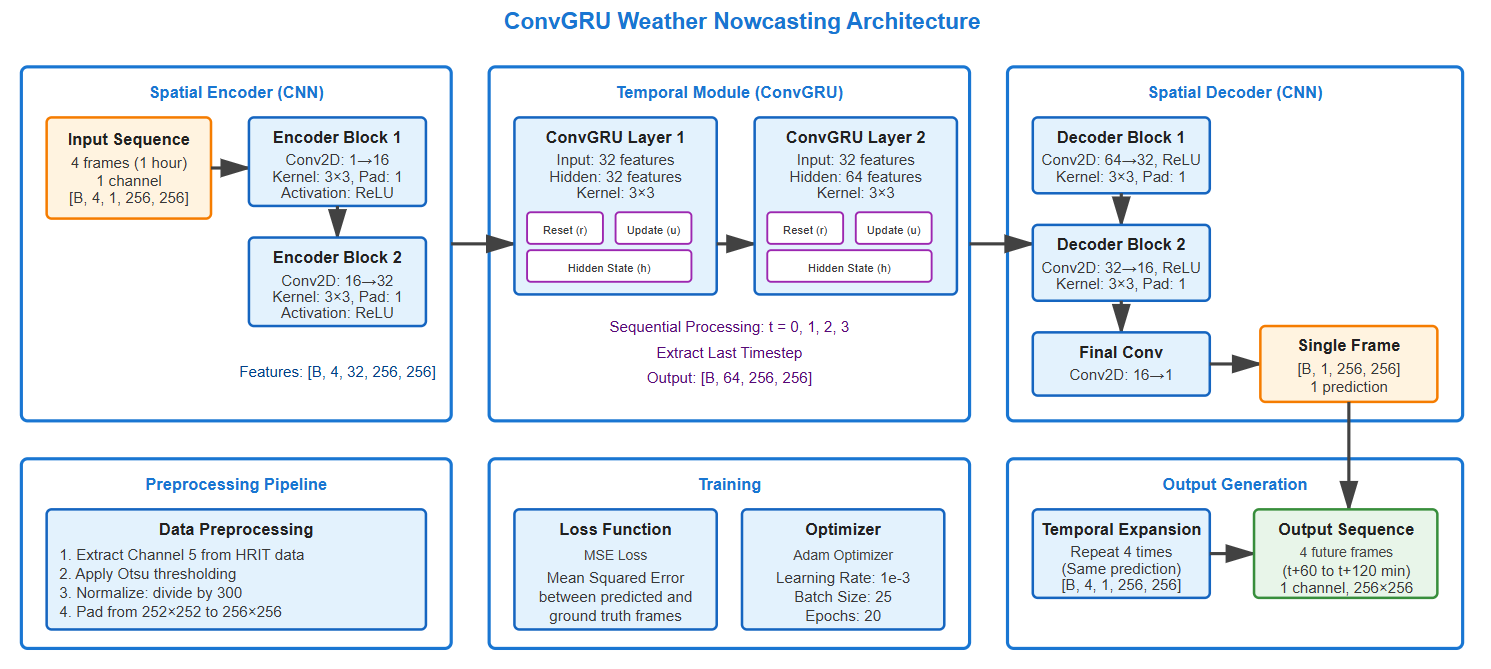}
    \caption{ConvGRU Weather Nowcasting Architecture.}
    \label{fig:convgru_arch}
\end{figure}

Our forecasting system used a ConvGRU based encoder-decoder architecture, Fig.~\ref{fig:convgru_arch} designed to capture both spatial cloud patterns and their temporal evolution. 
The encoder comprises two convolutional layers (1→16 and 16→32 filters, 3×3 kernels, ReLU activation). The temporal module consists of two stacked ConvGRU layers with hidden dimensions. Each ConvGRU cell updates hidden states through gated mechanisms, capturing spatiotemporal dependencies efficiently. The decoder reconstructs predicted brightness temperature fields through convolutional layers (64→32→16→1).

To achieve 4-hour-ahead forecasting (16 frames total), we trained four independent models with staggered temporal offsets of 1, 2, 3, and 4 hours respectively. Each model specializes in predicting 4 consecutive frames at its designated future horizon. This cascade approach offers several advantages over a single autoregressive model: (1) each model can specialize in the characteristic atmospheric dynamics at its temporal scale, (2) error accumulation from iterative forecasting is avoided, and (3) predictions at different horizons can be generated in parallel during inference.

We trained the model on the BOXI\_0015 dataset from the 2019 observation period. To extend its applicability beyond the training region, we adopted a transfer learning approach, fine-tuning the pretrained model to predict rainfall patterns in other geographical areas (ROXI regions) and for different years. This approach allowed the model to leverage previously learned spatial–temporal representations while adapting to new regional characteristics.

Each of the four models was trained independently using the Adam optimizer with learning rate $\alpha = 10^{-3}$. The loss function was Mean Squared Error (MSE) computed between predicted and ground-truth normalized brightness temperature fields. Training proceeded for 10 epochs with batch size 25.

For comparative analysis, we also implemented a ConvLSTM baseline architecture. The model consists of a two-layer ConvLSTM temporal module, with the same encoder–decoder CNN configuration and hyperparameters as the ConvGRU model. The ConvLSTM includes both hidden and cell states, enabling richer temporal memory at the expense of increased computational load. Despite its capacity, ConvLSTM models tend to overfit under limited training data and exhibit slower convergence due to their larger parameter space. 

A simple persistence model, which repeats the last observed frame as the forecast, is also included as a naive baseline to assess relative improvements.

Model outputs images of 252 x 252 size which are upscaled to 1512 x 1512 using bilinear interpolation to match OPERA radar resolution. The predicted values were converted using the empirically-derived transformation:
\[
\mathcal{R} = \alpha \times \max(0,\, 300 - \mathcal{T})^{\beta}
\]
where \(\mathcal{R}\) represents the rainfall rate (mm\,h$^{-1}$), and \(\mathcal{T}\) is the predicted brightness temperature (\(\mathcal{K}\)) and $\alpha$ and $\beta$ are empirically derived coefficients. This transformation, calibrated against OPERA radar data, captures the nonlinear relationship between cloud-top cooling and precipitation intensity (see \cite{goodman1994non, vicente1998operational}). 
The 16 transformed frames were averaged to obtain a 4-hour cumulative rainfall field, reducing prediction variance while capturing overall precipitation accumulation.

For cumulative rainfall predictions, spatially-averaged rainfall values were computed within predefined regions of interest by cropping predictions to specified coordinates. To address forecast uncertainty, deterministic predictions were converted to probabilistic outputs through cumulative distribution functions (CDFs). Threshold-probability pairs were generated with adaptive binning: rainfall values below 2 mm were assigned probability 1.0, while higher values were discretized into 0.5 mm intervals (probabilities 0.5, 0.75) and 2 mm intervals (Probability 1.0) to capture the full dynamic range of the rainfall distribution.

For the event prediction task of the Weather4Cast challenge, we adopt the same ConvGRU-based forecasting model used in the cumulative rainfall track. The model predicts future brightness temperature fields, which are subsequently transformed into rainfall rate using the same empirical HRIT-to-rainfall mapping employed in the cumulative task. This ensures architectural consistency across both tracks and provides a unified approach for handling spatiotemporal precipitation evolution.

The predicted rainfall-rate sequences (16 future frames) at 252×252 resolution were upsampled to 1512×1512 to match the native SEVIRI grid. These 3D volumes form the basis of the event detection pipeline. Rainfall events were identified using a three-dimensional connected-component labeling procedure applied to all voxels exceeding a 2 mm/h threshold. We use 18-connectivity to capture spatially and temporally coherent precipitation structures. For each detected event, descriptive attributes—including maximum intensity, duration, and spatial footprint—were computed. Bounding boxes and centroid positions from the central frame were used to estimate event location and approximate movement.
Finally, all events were converted into the standardized competition format by selecting the five most intense events per sequence and exporting them as the final event prediction CSVs.

\section{Results}
\begin{figure}[htb]
    \centering
    \includegraphics[width=1\textwidth]{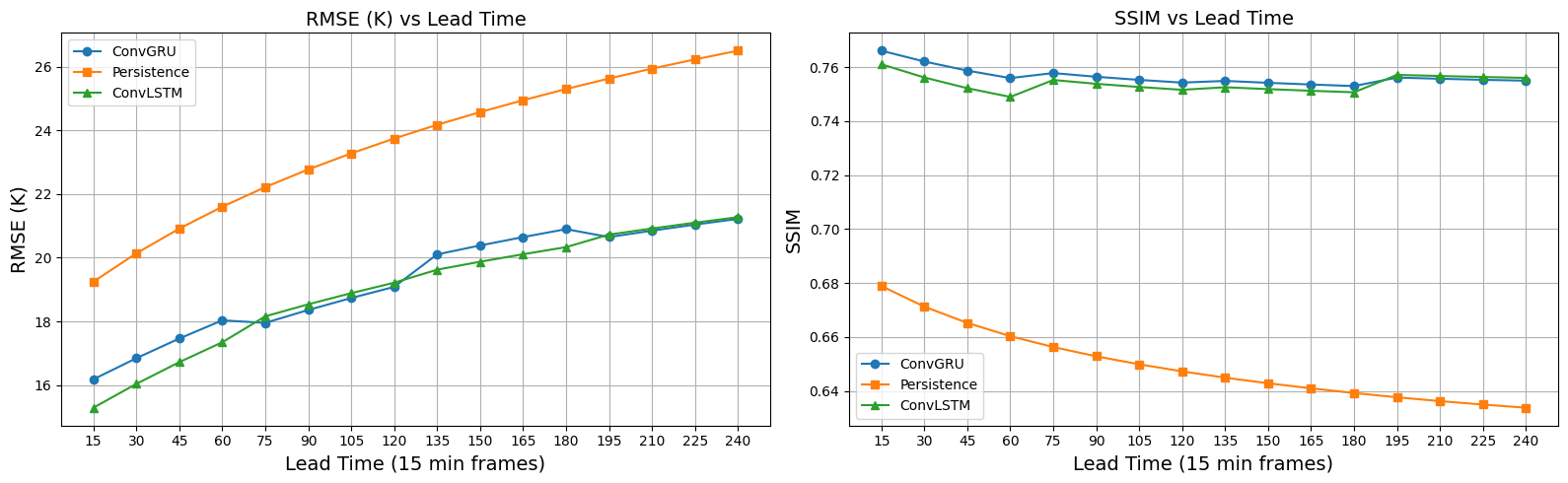}
    \caption{Comparison of ConvGRU, ConvLSTM, and Persistence models.}
    \label{fig:rmse_ssim_eval}
\end{figure}

To evaluate the model's performance we used Root Mean Squared Error (RMSE), Structural Similarity Index (SSIM) and Bias. RMSE helps us quantify pixel wise accuracy, while the SSIM measures perceptual similarity by comparing luminance, contrast and structure. We did a comparative study between ConvGRU, ConvLSTM and a persistence model. Figure~\ref{fig:rmse_ssim_eval} shows the temporal evolution of RMSE and SSIM with increasing forecast lead times (15-minute intervals). Both ConvLSTM and ConvGRU achieve substantially lower RMSE values and higher SSIM values than the persistence baseline across all lead times. The ConvLSTM showing slightly improved pixel-level accuracy at shorter horizons, whereas the ConvGRU marginally outperforms ConvLSTM across longer horizons, suggesting that its simplified gating mechanism enables more stable gradient propagation and better handling of extended temporal dependencies. 

\begin{figure}[htb]
    \centering
    \includegraphics[width=1\textwidth]{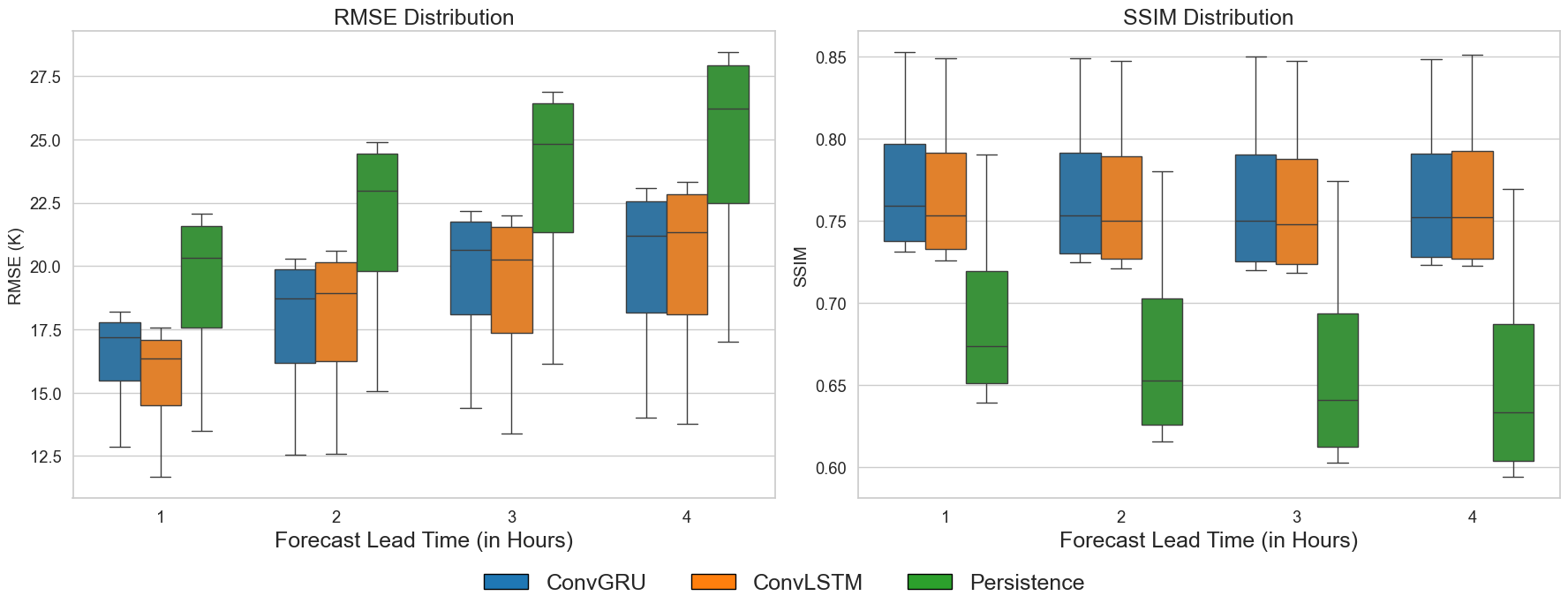}
    \caption{Model comparison across lead time.}
    \label{fig:box_plot}
\end{figure}
The box plots in Figure~\ref{fig:box_plot} represent the RMSE and SSIM distributions for the ROXI regions across 2019 and 2020 for all the three models and table~\ref{tab:model_comparison} reports the respective values for the training file (BOXI\_0015). It is evident that the RMSE and SSIM for the ROXI files closely correspond to those of the training set, suggesting that the transfer-learned model maintained robust performance and effectively generalized to unseen spatial domains.
\setlength{\tabcolsep}{4pt}
\begin{table}[htb]
\centering
\caption{Performance comparison of ConvGRU, ConvLSTM, and Persistence models for different forecast horizons.}
\begin{tabular}{lcccccc}
\toprule
\multirow{2}{*}{\textbf{Future Hours}} & 
\multicolumn{2}{c}{\textbf{ConvGRU}} & 
\multicolumn{2}{c}{\textbf{ConvLSTM}} & 
\multicolumn{2}{c}{\textbf{Persistence}} \\
\cmidrule(lr){2-3} \cmidrule(lr){4-5} \cmidrule(lr){6-7}
 & \textbf{RMSE ↓} & \textbf{SSIM ↑} & \textbf{RMSE ↓} & \textbf{SSIM ↑} & \textbf{RMSE ↓} & \textbf{SSIM ↑} \\
\midrule
1 & 17.12741 & 0.76069 & 16.34451 & 0.754578 & 20.47686 & 0.668903 \\
2 & 18.53167 & 0.75593 & 18.6989  & 0.753313 & 23.00733 & 0.651554 \\
3 & 20.51043 & 0.753894 & 19.98616 & 0.751569 & 24.75511 & 0.64199 \\
4 & 20.94211 & 0.755516 & 21.00648 & 0.756539 & 26.51112 & 0.633788 \\
\midrule
\textbf{Average} & \textbf{19.27791} & \textbf{0.756509} & \textbf{19.00901} & \textbf{0.754000} & \textbf{23.58056} & \textbf{0.649526} \\
\bottomrule
\end{tabular}
\label{tab:model_comparison}
\end{table}

To ensure that the evaluation reflects meteorologically meaningful conditions, we evaluate on a subset of frames in which more than 5\% of the spatial domain exhibits rainfall exceeding 5 mm. This filtering step removes near-empty or low-activity scenes that could otherwise inflate scores or mask model performance. We assess the HRIT-to-rainfall transformation against the OPERA ground truth using both continuous (RMSE and SSIM) and categorical metrics (POD - Probability of Detection, FAR - False Alarm Ratio and F1 score). As shown in Table~\ref{tab:rain_rates}, the transformation preserves key spatial precipitation patterns, achieves strong event-detection performance on active rainfall scenes, and provides a reliable quantitative baseline for evaluating learning-based approaches in the Weather4Cast cumulative rainfall track.

\begin{table}[htb]
\centering
\caption{Performance of the HRIT-to-rainfall transformation compared to OPERA ground truth.}
\begin{tabular}{lcccccc}
\hline
\textbf{Metrics} & \textbf{Values} \\ 
\hline
RMSE(mm) & 2.48 \\
SSIM & 0.747 \\
POD (at 0.5 mm) & 0.7713 \\
FAR (at 0.5 mm) & 0.3883 \\
F1 (at 0.5 mm) & 0.6822 \\
POD (at 1.0 mm) & 0.8018 \\
FAR (at 1.0 mm) & 0.5083 \\
F1 (at 1.0 mm) & 0.6095 \\
\hline
\end{tabular}
\label{tab:rain_rates}
\end{table}

Our model achieved strong generalization on the cumulative rainfall task achieving a score of 3.37 and scored a similar result as the baseline on the event prediction task, ranking second on the official leaderboard of the Weather4Cast 2025 Challenge. These outcomes indicate that the proposed architecture not only captures spatiotemporal patterns effectively but also generalizes well across diverse meteorological conditions. The qualitative findings presented above are further supported by the Continuous Ranked Probability Score (CRPS) \cite{zamo_naveau:2018}, which served as the primary evaluation metric for ranking submissions in the Weather4Cast 2025 competition.

\section{Conclusion}
In this work, we investigated spatiotemporal deep learning architectures for precipitation forecasting, focusing on the comparative performance of ConvLSTM and ConvGRU models within the Weather4Cast 2025 framework. Our experiments demonstrate that while both architectures effectively capture local and temporal rainfall dynamics, the ConvGRU architecture proved more effective for cumulative rainfall forecasting, where faster convergence and reduced parameter complexity allowed better generalization across regions. Whereas, the ConvLSTM model exhibited stronger performance for event-based prediction, benefiting from its higher temporal sensitivity and ability to retain long-term dependencies. Using a transfer learning approach, the model trained on the BOXI\_0015 (2019) dataset was successfully adapted to unseen ROXI regions for multiple years, indicating its capacity for spatial transferability.

\vskip 0.2in
\bibliography{sample}

@Article{deshpande:2024,
  author       = {Atharva Deshpande and Kaushik Gopalan and Jeet Shah and Hrishikesh Simu},
  title        = {A Conditional Generative Adversarial Network Model for the Weather4Cast 2024 Challenge},
  journal      = {arXiv preprint arXiv:2412.00451},
  year         = {2024},
  url          = {https://arxiv.org/abs/2412.00451}
}

@Article{otsu:1975,
  author       = {Otsu, Nobuyuki},
  title        = {A Threshold Selection Method from Gray-Level Histograms},
  journal      = {Automatica},
  volume       = {11},
  pages        = {23--27},
  year         = {1975}
}

@Article{zamo_naveau:2018,
  author       = {Michaël Zamo and Philippe Naveau},
  title        = {Estimation of the Continuous Ranked Probability Score with Limited Information and Applications to Ensemble Weather Forecasts},
  journal      = {Mathematical Geosciences},
  year         = {2018},
  volume       = {50},
  number       = {2},
  pages        = {209--234},
  doi          = {10.1007/s11004-017-9709-7}
}

@Article{reichstein:2019,
  author       = {Reichstein, Markus and Camps-Valls, Gustau and Stevens, Bjorn and Jung, Martin and Denzler, Joachim and Carvalhais, Nuno},
  title        = {Deep learning and process understanding for data-driven Earth system science},
  journal      = {Nature},
  volume       = {566},
  pages        = {195--204},
  year         = {2019},
  publisher    = {Nature Publishing Group}
}

@Article{shi:2015,
  author       = {Shi, Xingjian and Chen, Zhourong and Wang, Hao and Yeung, Dit-Yan and Wong, Wai-kin and Woo, Wang-chun},
  title        = {Convolutional LSTM Network: A Machine Learning Approach for Precipitation Nowcasting},
  journal      = {Advances in Neural Information Processing Systems (NeurIPS)},
  year         = {2015},
  url          = {https://papers.nips.cc/paper_files/paper/2015/hash/07563a3fe3bbe7e3ba84431ad9d055af-Abstract.html}
}

@Article{rasp_dueben_scher_weyn_mouatadid_thuerey:2020,
  author       = {Stephan Rasp and Peter D. Dueben and Sebastian Scher and Jonathan A. Weyn and Soukayna Mouatadid and Nils Thuerey},
  title        = {WeatherBench: A Benchmark Data Set for Data-Driven Weather Forecasting},
  journal      = {Journal of Advances in Modeling Earth Systems},
  year         = {2020},
  volume       = {12},
  number       = {11},
  pages        = {e2020MS002203},
  doi          = {10.1029/2020MS002203},
  url          = {https://doi.org/10.1029/2020MS002203}
}

@Article{bauer_thorpe_brunet:2015,
  author       = {Peter Bauer and Alan Thorpe and Gilbert Brunet},
  title        = {The Quiet Revolution of Numerical Weather Prediction},
  journal      = {Nature},
  year         = {2015},
  volume       = {525},
  number       = {7567},
  pages        = {47--55},
  doi          = {10.1038/nature14956},
  url          = {https://doi.org/10.1038/nature14956}
}

@Article{schultz_betancourt_gong_kleinert_langguth_leufen:2019,
  author       = {Michaël G. Schultz and C. Betancourt and B. Gong and F. Kleinert and M. Langguth and L. H. Leufen},
  title        = {Can deep learning beat numerical weather prediction?},
  journal      = {Bulletin of the American Meteorological Society},
  year         = {2019},
  volume       = {100},
  number       = {7},
  pages        = {1337--1354}
}

@InProceedings{veillette_samsi_mattioli:2020,
  author       = {Mark S. Veillette and Siddharth Samsi and Christopher J. Mattioli},
  title        = {SEVIR: A Storm Event Imagery Dataset for Deep Learning Applications in Radar and Satellite Meteorology},
  booktitle    = {Advances in Neural Information Processing Systems 33},
  year         = {2020},
  pages        = {22009--22019},
  url          = {https://proceedings.neurips.cc/paper/2020/hash/fa78a16157fed00d7a80515818432169-Abstract.html}
}

@Article{wang_wu_zhang_gao_wang_yu_long:2022,
  author       = {Yunbo Wang and Haixu Wu and Jianjin Zhang and Zhifeng Gao and Jianmin Wang and Philip Yu and Mingsheng Long},
  title        = {PredRNN: A Recurrent Neural Network for Spatiotemporal Predictive Learning},
  journal      = {IEEE Transactions on Pattern Analysis and Machine Intelligence},
  year         = {2022},
  note         = {to appear}
}

@Article{woo_wong:2017,
  author       = {Wang-chun Woo and Wai-kin Wong},
  title        = {Operational Application of Optical Flow Techniques to Radar-Based Rainfall Nowcasting},
  journal      = {Atmosphere},
  year         = {2017},
  volume       = {8},
  number       = {3},
  pages        = {48},
  doi          = {10.3390/atmos8030048},
  url          = {https://doi.org/10.3390/atmos8030048}
}

@inproceedings{gruca2023weather4cast,
  title={Weather4cast at neurips 2022: Super-resolution rain movie prediction under spatio-temporal shifts},
  author={Gruca, Aleksandra and Serva, Federico and Lliso, Lloren{\c{c}} and R{\'\i}podas, Pilar and Calbet, Xavier and Herruzo, Pedro and Pihrt, Ji{\v{r}}{\'\i} and Raevskyi, Rudolf and {\v{S}}im{\'a}nek, Petr and Choma, Matej and others},
  booktitle={NeurIPS 2022 competition track},
  pages={292--313},
  year={2023},
  organization={PMLR}
}

@article{leinonen2021spatiotemporal,
  title={Spatiotemporal weather data predictions with shortcut recurrent-convolutional networks: A solution for the Weather4cast challenge},
  author={Leinonen, Jussi},
  journal={arXiv preprint arXiv:2111.02121},
  year={2021}
}

@article{goodman1994non,
  title={A non-linear algorithm for estimating three-hourly rain rates over Amazonia from GOES/VISSR observations},
  author={Goodman, Brian and Martin, David W and Menzel, W Paul and Cutrim, Elen C},
  journal={Remote Sensing Reviews},
  volume={10},
  number={1-3},
  pages={169--177},
  year={1994},
  publisher={Taylor \& Francis}
}

@article{vicente1998operational,
  title={The operational GOES infrared rainfall estimation technique},
  author={Vicente, Gilberto A and Scofield, Roderick A and Menzel, W Paul},
  journal={Bulletin of the American Meteorological Society},
  volume={79},
  number={9},
  pages={1883--1898},
  year={1998},
  publisher={American Meteorological Society}
}

\end{document}